\RequirePackage[svgnames]{xcolor}

\documentclass[11pt,letterpaper]{arxiv/mystyle}

\usepackage[all]{hypcap}
\usepackage[svgnames]{xcolor}
\usepackage[comma,authoryear,compress]{natbib}
\usepackage{hyperref}[citecolor=lightblue]

\hypersetup{
    colorlinks = true,
    citecolor = {YaleBlue},
}

\usepackage{algorithm}
\usepackage{algorithmicx}
\usepackage{algpseudocode}
\usepackage{microtype}
\usepackage{graphicx}
\expandafter\def\csname ver@subfig.sty\endcsname{}
\usepackage{booktabs} %
\usepackage{xspace}
\usepackage{float}
\usepackage{bigstrut}

\usepackage{amsmath}
\usepackage{amssymb}
\usepackage{mathtools}
\usepackage{amsthm}
\usepackage{mathrsfs}
\usepackage{nicefrac}
\usepackage{dsfont}
\usepackage{enumitem}
\usepackage{subcaption}
\usepackage{graphicx,subfig}
\usepackage{cleveref}
\usepackage{bxcoloremoji}

\usepackage{float}

\usepackage[utf8]{inputenc} %
\usepackage[T1]{fontenc}    %
\usepackage{hyperref}       %
\usepackage{url}            %
\usepackage{booktabs}       %
\usepackage{amsfonts}       %
\usepackage{nicefrac}       %
\usepackage{microtype}      %
\usepackage{graphicx}
\usepackage{subcaption}
\usepackage{amssymb}
\usepackage{fdsymbol}
\usepackage{wrapfig}
\usepackage{lipsum}
\usepackage{enumitem}
\usepackage{stackengine}
\usepackage[font=small,labelfont=bf]{caption}
\usepackage{color}
\usepackage{adjustbox}
\usepackage{multirow}

\usepackage{rotating}
\usepackage{makecell}

\newcommand{\x}{\mathbf{x}}

\definecolor{blanchedalmond}{rgb}{1.0, 0.92, 0.8}
\definecolor{carmine}{rgb}{0.59, 0.0, 0.09}
\definecolor{lightblue}{rgb}{0.22,0.45,0.70}%

\renewcommand{\mathbf}{\boldsymbol}

\makeatletter
\def\Ddots{\mathinner{\mkern1mu\raise\p@
\vbox{\kern7\p@\hbox{.}}\mkern2mu
\raise4\p@\hbox{.}\mkern2mu\raise7\p@\hbox{.}\mkern1mu}}
\makeatother

\definecolor{amaranth}{rgb}{0.9, 0.17, 0.31}
\definecolor{antiquebrass}{rgb}{0.8, 0.58, 0.46}
\definecolor{antiquefuchsia}{rgb}{0.57, 0.36, 0.51}
\definecolor{chromeyellow}{rgb}{0.31, 0.47, 0.26}

\newtcolorbox{AIbox}[2][]{aibox,title=#2,#1}
\definecolor{lightblue}{rgb}{0.22,0.45,0.70}%
\definecolor{Gray}{gray}{0.95}
\definecolor{Cornsilk}{rgb}{1.0, 0.97, 0.86}

\newcommand{\method}{VCSD\xspace} 

\usepackage{amsmath}

\usepackage[all]{hypcap}
\definecolor{gainGreen}{HTML}{3B9B72}
\newcommand{\gain}[1]{\textcolor{gainGreen}{\textbf{(+#1)}}}

\title{Visual Contrastive Self-Distillation}

\runningtitle{Visual Contrastive Self-Distillation}

\author{
  Yijun Liang\textsuperscript{1},
  Yunjie Tian\textsuperscript{5},
  Yijiang Li\textsuperscript{2},
  Yuqi Jia\textsuperscript{3}, 
  Kaishen Wang\textsuperscript{1},
  Tianyi Zhou\textsuperscript{4},
  Di Fu\textsuperscript{5}\\
  {\normalfont\fontsize{10}{12}\selectfont
  \textsuperscript{1}University of Maryland, College Park
  \quad
  \textsuperscript{2}UC San Diego
  \quad
  \textsuperscript{3}Duke University
  \quad
  \textsuperscript{4}MBZUAI}
  \quad
  \textsuperscript{5}ByteDance}

\begin{document}

\begin{abstract}

On-policy self-distillation (OPSD) removes the external teacher required by on-policy distillation (OPD), but still needs asymmetric information between teacher and student so that the self-teacher provides a stronger learning signal.
Existing methods create this asymmetry through privileged answers or visual evidence. We ask whether both can be removed, yielding a simpler form of OPSD driven purely by input conditioning.
We propose \textbf{Visual Contrastive Self-Distillation}, namely \textbf{\method}, which converts image-content removal into an on-policy self-distillation signal.
At each student-generated response prefix, the EMA teacher produces two next-token distributions under the same prompt and prefix: one conditioned on the original image and the other on a content-erased control.
Their token-wise log-probability difference highlights candidates whose likelihood is specifically increased by the instance-level visual content.
We use this contrast to sharpen the teacher's original-image distribution within its plausible support, and distill the resulting full-distribution target into the student.
On ViRL39K, \method consistently outperforms matched OPSD across Qwen3-VL and Qwen3.5 models. It improves the seven-benchmark aggregate from $72.51\%$ to $76.26\%$ on Qwen3-VL-8B and from $74.97\%$ to $79.24\%$ on Qwen3.5-9B, with gains at all six evaluated scales.
\method requires no external teacher, privileged answers, visual evidence signals, reasoning traces, or additional inference-time cost.

\mbox{Project page:} \url{https://joliang17.github.io/VisualCSD/}
\end{abstract}

\maketitle

\vspace{3mm}

\section{Introduction}
\label{sec:intro}

On-policy distillation (OPD)~\citep{agarwal2024policy,lu2025onpolicydistillation,li2026rethinking} trains a student on prefixes sampled from its own policy while using an external teacher for dense token-level supervision.
This aligns training with the student's inference-time trajectories, but typically requires a stronger teacher, increasing post-training cost and complexity~\citep{liu2026visual,yoon2026decomposed}.
On-policy self-distillation (OPSD) removes the external teacher by deriving targets from the same model, often through an exponential moving average (EMA) teacher.
However, when the student and self-teacher receive the same information at the same prefix, the target may add little beyond the student's current prediction.
Thus, on-policy sampling determines \emph{where} learning occurs, but not \emph{what} additional supervision the self-teacher provides.
Effective OPSD still requires a teacher--student asymmetry that makes the self-teacher more informative than the student.

\begin{figure}[t]
  \centering
  \includegraphics[width=\linewidth]{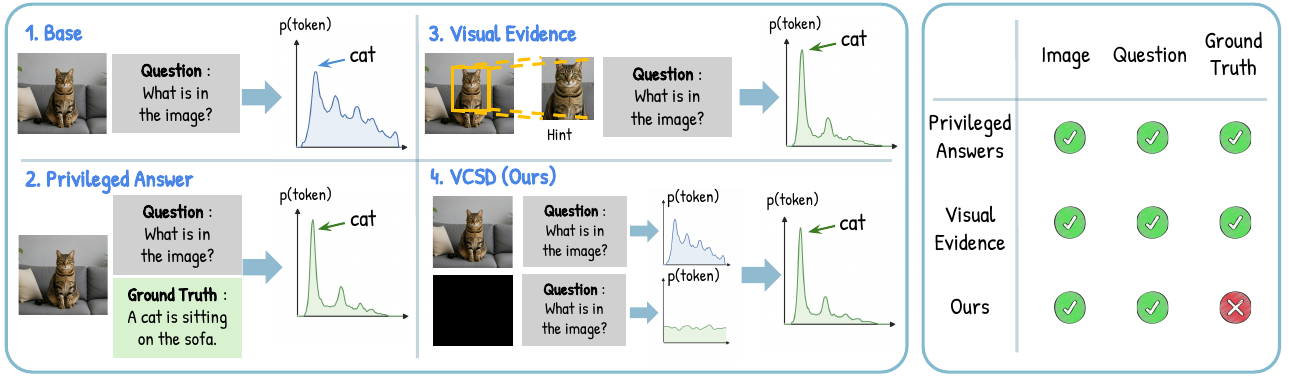}
  \caption{\textbf{Comparison of different sources of target asymmetry.}
Existing methods sharpen the teacher distribution using privileged answers or visual evidence.
Our method instead contrasts teacher predictions under the original image and a content-erased control, constructing a visually informed target using only the image and question.}
\label{fig:motivition}
\vspace{-0.4cm}
\end{figure}

Existing methods construct this asymmetry through auxiliary information that is available to the self-teacher but not to the student, as illustrated in Figure~\ref{fig:motivition}.
For language reasoning tasks, the teacher may be conditioned on privileged answers or reasoning traces~\citep{zhao2026self}, expert demonstrations~\citep{shenfeld2026self}, or rich textual feedback~\citep{hubotter2026reinforcement}.
For vision-language tasks, the teacher can receive visual evidence signals, such as crops, regions, or other visual conditions that expose the relevant image evidence more directly~\citep{yuan2026vision,sun2026v,tian2026vicur}.
These approaches can produce effective supervision, but depend on task-specific information or additional processing pipelines.
These language-side signals are not always available, while visual evidence inputs can require annotations, localization models, or custom image transformations.

This motivates the central question of this work:
\textit{\textbf{Can the asymmetry required by on-policy self-distillation be constructed purely from input conditioning, without auxiliary information?}}
We answer this question by treating a content-erased image not as an alternative teacher input to imitate, but as a controlled reference.
At each student-generated response prefix, the same EMA teacher is evaluated twice under the same prompt and prefix: once with the original image and once with a content-erased control.
Since only the visual condition changes, their token-wise log-probability difference captures how each candidate's likelihood changes when the instance-specific visual content is removed, as illustrated in Figure~\ref{fig:motivition}.
This conditioning contrast provides the target asymmetry needed for self-distillation.

The contrast alone, however, does not define a reliable distillation target, since a token may exhibit a large relative change while remaining unlikely under the original image.
We therefore assign complementary roles to the two teacher predictions: the original-image distribution identifies plausible candidates, while the conditioning contrast sharpens their relative preferences according to their dependence on visual content.
Together, they yield a visually informed full-distribution target for distilling the student on its own trajectory.

Building on this idea, we propose \textbf{Visual Contrastive Self-Distillation}, namely \method.
Given a student-generated prefix $y_{<t}$, we define
\vspace{-0.3cm}

\begin{equation}
\begin{aligned}
\Delta_t(v)
&=
\log p_{\phi}(v \mid P,I,y_{<t})
-
\log p_{\phi}(v \mid P,I_{\mathrm{ctrl}},y_{<t}), \\
q_t^*(v)
&\propto
p_{\phi}(v \mid P,I,y_{<t})
\exp\!\left(\alpha\Delta_t(v)\right),
\qquad v \in S_t,
\end{aligned}
\end{equation}
\vspace{-0.3cm}

where $S_t$ restricts the target to candidates that are plausible under the original image.
The resulting full-distribution target is distilled into the student through forward KL along its on-policy trajectory.
Here, ``contrastive'' refers to the contrast between conditional token distributions rather than an embedding-level contrastive loss.

We evaluate \method on ViRL39K across Qwen3-VL and Qwen3.5 models from 2B to 9B using seven vision-language benchmarks.
It improves the aggregate score from $72.51\%$ to $76.26\%$ on Qwen3-VL-8B and from $74.97\%$ to $79.24\%$ on Qwen3.5-9B, with gains at every evaluated scale.
It also consistently improves Qwen3.5 models by $2.8\%$ to $4.3\%$ over their corresponding base models.
Across model families and scales, \method outperforms both the base models and matched OPSD baselines.
Moreover, \method requires no external teacher, privileged answer, reasoning trace, evidence-focused crop, or external verifier.

Our contributions are threefold:
\begin{itemize}
    \item We show that matched input conditioning can provide the target asymmetry required for OPSD without privileged answers or visual evidence signals.

    \item We propose \method, which uses the contrast between original-image and content-erased teacher predictions to sharpen a plausible full-distribution target.

    \item We demonstrate consistent improvements across Qwen3-VL and Qwen3.5 models from 2B to 9B on seven vision-language benchmarks.
\end{itemize}


\section{Related Work}
\label{sec:related}

\textbf{Target asymmetry in on-policy self-distillation.}
On-policy self-distillation builds its target from the same underlying model, requiring an asymmetry that makes it informative relative to the student.
Existing methods create this asymmetry with reference or self-generated solutions~\citep{zhao2026self,li2026policy}, evidence-centered or augmented visual views~\citep{yuan2026vision,li2026self,wang2026look}, or paired evidence for trajectory selection~\citep{sun2026v}.
Other supervision regimes use verifiable rewards~\citep{shao2024deepseekmath,yang2026self} or separately trained teachers, sometimes with visual weighting or gradient steering~\citep{liu2026visual,bousselham2026vold,yoon2026decomposed}.
\method keeps the student conditioned on the original image and constructs its target from the same teacher's predictions under original and content-erased image conditions.
The content-erased prediction supplies a contrastive reference, while the original-image prediction anchors the target to plausible candidates.
This targets the tendency of VLMs to favor linguistic priors over fine-grained image evidence~\citep{guan2024hallusionbench}.
At inference time, related methods address this dependency through contrastive decoding across models, visual conditions, or internal representations~\citep{li2023contrastive,leng2024mitigating}, or through layer-wise activation refinement~\citep{wang2025damo}.
\method distills the visual contrast into OPSD without verifiable rewards, external teachers, or extra inference-time computation.

\textbf{Relative predictive distributions under paired conditions.}
Differences between predictive distributions have been used to expose information that is not apparent from either prediction in isolation.
Contrastive decoding compares expert and amateur models, or original and degraded visual conditions, to modify token selection at inference~\citep{li2023contrastive,leng2024mitigating}.
Beyond decoding, paired model behaviors can also define training signals: MARGO uses non-thinking rollouts as same-model references when estimating the advantage of explicit reasoning~\citep{wang2026mitigating}.
Related policy-transfer methods compare model checkpoints: weak-to-strong preference optimization transfers changes introduced by alignment~\citep{zhu2025weak}, while Direct-OPD transfers an RL-induced policy difference to a student's on-policy states~\citep{feng2026weak}.
\method considers a same-model, paired-input setting in which the target parameters and response prefix are held fixed while only the visual condition changes.
The resulting vocabulary-level contrast shapes a detached token distribution that is distilled by forward KL.
The real-image prediction supplies the plausibility anchor, and the content-erased prediction supplies the reference used to adjust its relative token preferences. \looseness-1

\section{Method}
\label{sec:method}

\begin{figure}[t]
  \centering
  \includegraphics[width=\linewidth]{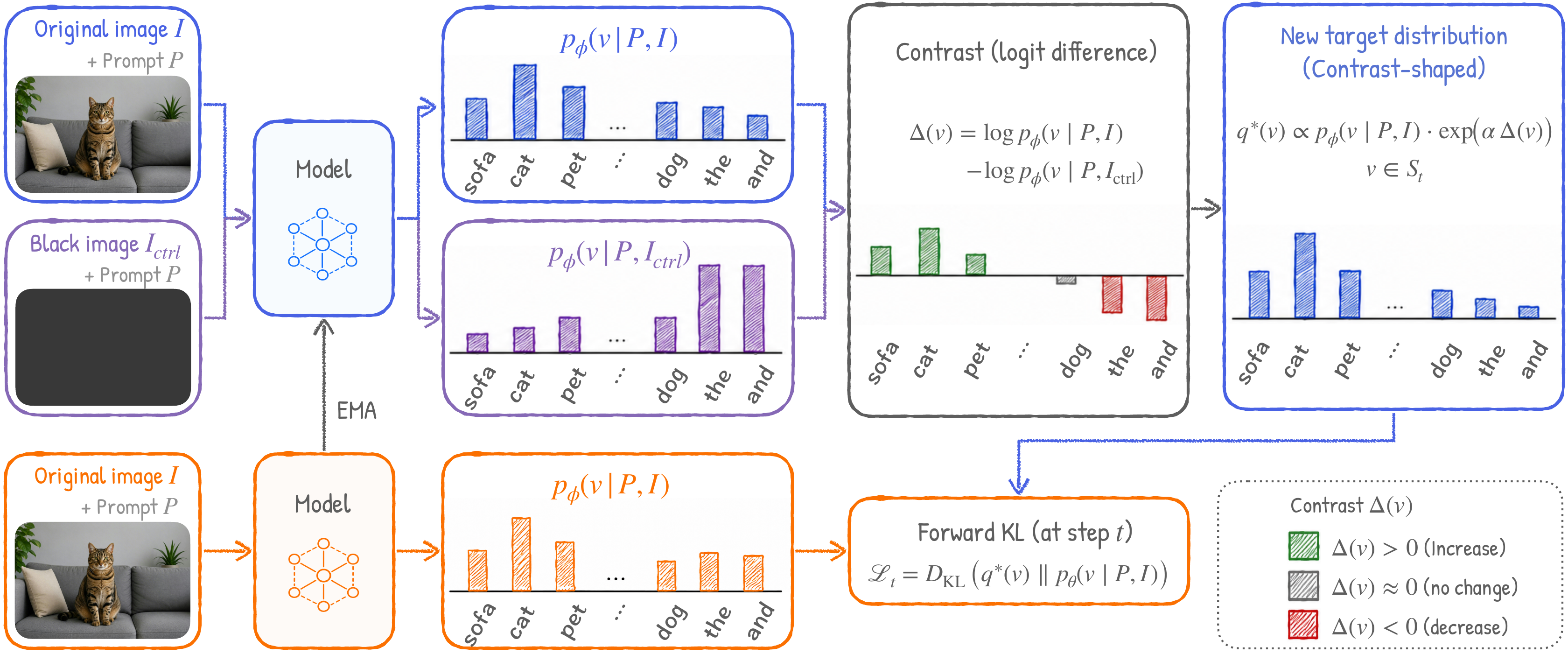}
  \caption{\textbf{Overview of \method.}
  At each student-generated prefix, the EMA teacher contrasts next-token distributions under the original image and a content-erased control.
  The contrast sharpens the plausible original-image distribution to form the target, which is distilled into the student via forward KL.
  The response prefix is omitted for clarity.}
  \label{fig:framework}
\vspace{-0.4cm}
\end{figure}

We propose \method, which constructs the target asymmetry required for OPSD from matched visual conditioning.
As shown in Figure~\ref{fig:framework}, the student generates an on-policy response under the original image, while the EMA teacher evaluates each response prefix under both the original image and a content-erased control.
Their token-wise log-probability contrast sharpens the original-image distribution within its plausible support, yielding a full-distribution target for self-distillation.

\subsection{On-Policy Self-Distillation Setup}
\label{sec:opsd-setup}

Let \(\mathcal{D}=\{(P_i,I_i)\}_{i=1}^{N}\) denote a collection of prompt--image pairs, where \(P\) is the prompt and \(I\) is the original image.
We denote the trainable student by \(\pi_\theta\) and its EMA teacher by \(\pi_\phi\).
The teacher serves as a stop-gradient target; its parameters receive no gradients.

For each \((P,I)\sim\mathcal{D}\), the student samples an on-policy response

\vspace{-0.3cm}
\begin{equation}
  y \sim \pi_\theta(\cdot\mid P,I).
  \label{eq:on-policy-rollout}
\end{equation}

\vspace{-0.3cm}

At generation step \(t\), both the student and teacher are evaluated along the same student-generated prefix \(y_{<t}\).
The student next-token distribution is

\vspace{-0.3cm}
\begin{equation}
  p_{\theta,t}(v)
  =
  \pi_\theta(v\mid P,I,y_{<t}),
  \qquad v\in\mathcal{V},
  \label{eq:student-distribution}
\end{equation}

\vspace{-0.3cm}
where \(\mathcal V\) is the vocabulary.

The rollout is on-policy because the student determines every prefix at which supervision is applied.
However, on-policy sampling alone does not specify how the teacher target should become more informative than \(p_{\theta,t}\).
Instead of providing the teacher with a privileged answer or an evidence-focused crop, \method constructs this target asymmetry by comparing the teacher under two matched visual conditions.

\subsection{Visual Conditioning Contrast}
\label{sec:visual-conditioning-contrast}

We construct a content-erased control \(I_{\mathrm{ctrl}}=\mathcal{C}(I)\), where \(\mathcal{C}(I)\) is a same-size black RGB image.
The control preserves the image resolution, multimodal input interface, preprocessing path, and visual-token count, while removing the instance-specific content of \(I\).

At each fixed prefix \(y_{<t}\), the EMA teacher produces two next-token distributions:

\vspace{-0.3cm}
\begin{equation}
  p_{\phi,t}^{I}(v)
  =
  \pi_\phi(v\mid P,I,y_{<t}),
  \label{eq:teacher-original}
  \quad
  p_{\phi,t}^{0}(v)
  =
  \pi_\phi(v\mid P,I_{\mathrm{ctrl}},y_{<t}).
\end{equation}

The prompt, response prefix, model parameters, and multimodal computation path are shared between the two evaluations.
They differ only in whether the instance-specific visual content is present.

We measure the change in token preference through the vocabulary-level log-probability contrast

\vspace{-0.3cm}
\begin{equation}
  \Delta_t(v)
  =
  \log p_{\phi,t}^{I}(v)
  -
  \log p_{\phi,t}^{0}(v).
  \label{eq:conditioning-contrast}
\end{equation}
\vspace{-0.3cm}

A positive \(\Delta_t(v)\) indicates that token \(v\) receives greater support under the original image than under the content-erased control.
Conversely, a negative value indicates that its probability increases when the original visual content is removed.
Thus, \(\Delta_t\) describes how the teacher's next-token preferences change in response to the instance-specific visual content.

Importantly, the control distribution is used only as a reference.
It is neither treated as a teacher target nor directly distilled into the student.

\subsection{Contrast-Shaped Teacher Target}
\label{sec:contrast-shaped-target}

The conditioning contrast provides a direction for modifying the teacher target, but it is not sufficient on its own.
A token may exhibit a large relative change between the two conditions while remaining highly unlikely under the original image.
We therefore use the original-image teacher distribution as a plausibility anchor and apply contrastive shaping only to candidates that it considers sufficiently likely.

Specifically, we define the relative plausibility support

\vspace{-0.3cm}
\begin{equation}
  \mathcal{S}_t(\beta)
  =
  \left\{
    v\in\mathcal{V}
    \;:\;
    p_{\phi,t}^{I}(v)
    \ge
    \beta
    \max_{u\in\mathcal{V}}
    p_{\phi,t}^{I}(u)
  \right\},
  \label{eq:plausibility-support}
\end{equation}

\vspace{-0.3cm}
where \(\beta\in[0,1]\) controls the support threshold.
This relative-threshold construction follows the plausibility restriction commonly used in contrastive decoding~\citep{li2023contrastive,leng2024mitigating}.

We then construct the contrast-shaped teacher target as

\vspace{-0.3cm}
\begin{equation}
  q_t^\star(v)
  =
  \frac{
    \mathbf{1}
    \left[
      v\in\mathcal{S}_t(\beta)
    \right]
    p_{\phi,t}^{I}(v)
    \exp\!\left(
      \alpha\widetilde{\Delta}_t(v)
    \right)
  }{
    \displaystyle
    \sum_{u\in\mathcal{S}_t(\beta)}
    p_{\phi,t}^{I}(u)
    \exp\!\left(
      \alpha\widetilde{\Delta}_t(u)
    \right)
  },
  \label{eq:contrast-shaped-target}
\end{equation}

\vspace{-0.3cm}
where \(\alpha\geq0\) controls the strength of contrastive shaping.
Here, \(\widetilde{\Delta}_t\) equals \(\Delta_t\) except that the contrast values of designated sequence-termination tokens are set to zero.

Equation~\ref{eq:contrast-shaped-target} assigns complementary roles to the two signals.
The original-image distribution \(p_{\phi,t}^{I}\) determines the initial probability and the admissible candidate set, while \(\widetilde{\Delta}_t\) adjusts the relative probabilities according to how strongly each candidate is favored by the original visual content.
Within the support, the corresponding unnormalized log score is

\vspace{-0.3cm}
\begin{equation}
  \log p_{\phi,t}^{I}(v)
  +
  \alpha\widetilde{\Delta}_t(v)
  =
  (1+\alpha)\log p_{\phi,t}^{I}(v)
  -
  \alpha\log p_{\phi,t}^{0}(v),
  \label{eq:equivalent-target-score}
\end{equation}

\vspace{-0.3cm}
for non-termination tokens.
When \(\alpha=0\), the target reduces to the original-image teacher distribution renormalized over \(\mathcal S_t(\beta)\).
When \(\beta=0\), the target spans the full vocabulary.

In \method, ``contrastive'' therefore refers to the contrast between two conditional token distributions, rather than positive and negative embedding pairs.

\subsection{On-Policy Distillation and EMA Update}
\label{sec:on-policy-distillation}

The contrast-shaped target is distilled into the student at every position of the student-generated response using full-distribution forward KL:

\vspace{-0.3cm}
\begin{equation}
  \mathcal{L}_{\method}
  =
  T_{\mathrm{KD}}^2
  \,
  \mathbb{E}_{\substack{
    (P,I)\sim\mathcal{D}\\
    y\sim\pi_\theta(\cdot\mid P,I)
  }}
  \left[
    \frac{1}{|y|}
    \sum_{t=1}^{|y|}
    D_{\mathrm{KL}}
    \left(
      \operatorname{sg}
      \left[
        q_t^\star
      \right]
      \,\middle\|\,
      p_{\theta,t}
    \right)
  \right],
  \label{eq:vcsd-loss}
\end{equation}

\vspace{-0.3cm}
where \(\operatorname{sg}[\cdot]\) denotes stop-gradient.
All teacher and student distributions used for target construction and distillation are evaluated with the same distillation temperature \(T_{\mathrm{KD}}\); its dependence is omitted from the notation for clarity.
The factor \(T_{\mathrm{KD}}^2\) follows standard temperature-scaled knowledge distillation.

Gradients pass only through the student distribution \(p_{\theta,t}\).
The two teacher evaluations are performed on the same fixed student-generated prefix and do not produce separate response trajectories.
After each student update, the teacher parameters are updated by \(\phi
  \leftarrow
  \mu\phi+(1-\mu)\theta\), where \(\mu\in[0,1)\) is the EMA decay coefficient.

Training requires only the original prompt--image pair and the student's own on-policy response.
It does not consume an external teacher, privileged answer, reasoning trace, evidence-focused crop, external verifier, or verifiable reward.
At inference time, only the updated student is retained; neither the EMA teacher nor the content-erased control introduces an additional inference-time branch.

\subsection{Policy-Improvement Interpretation}
The conditional log-ratio in Equation~\ref{eq:conditioning-contrast} acts as a dense token-level reward for support attributable to instance-specific visual content.
Under this view, the contrast-shaped target is the unique optimizer of a one-step KL-regularized policy-improvement objective: the original-image prediction supplies the reference policy, the conditioning contrast supplies the reward, and the plausibility set constrains the update to candidates already supported by the observed input.
This connects input-conditioned self-distillation to conservative policy improvement without requiring an external reward model; Appendix~\ref{sec:theoretical-perspective} gives the complete formulation and its conditional-information interpretation.


\section{Experiments}
\label{sec:experiments}

\subsection{Setup}
\label{sec:setup}
\textbf{Models and training data.}
We evaluate \method on Qwen3-VL~\citep{bai2025qwen3} at 2B / 4B / 8B and on Qwen3.5~\citep{qwen35} at 2B / 4B / 9B.
The main experiments train all post-training methods on ViRL39K single-image dataset~\citep{wang2026vl}.
Section~\ref{sec:vision-opd-generalization} additionally retrains \method on Vision-OPD6K training set~\citep{yuan2026vision} to evaluate generalization across training data. \looseness-1

\textbf{Baselines.}
Each model block compares the unmodified base model, the published answer-hint OPSD method~\citep{zhao2026self}, and \method.
GRPO uses rule-based verifiable rewards and its own tuned hyperparameters, while SFT uses cached non-thinking teacher responses.
All arms match optimization steps and prompt budget.
The OPSD target model is conditioned on the reference answer while the student receives only the original question and image.
\method retains the on-policy rollout structure but constructs its target from the paired original and content-erased image conditions without consuming the answer.
Implementation details and training hyperparameters are provided in Appendix~\ref{sec:appendix-repro}. \looseness-1

\textbf{Benchmarks and aggregate metric.}
We evaluate general visual perception with BLINK~\citep{fu2024blink} and MMStar~\citep{chen2024we}, visual mathematics with MathVista~\citep{lu2024mathvista}, fine-grained and high-resolution perception with V$^*$Bench~\citep{vstar} and HRBench4K/8K~\citep{wang2025divide}, and hallucination with HallusionBench~\citep{guan2024hallusionbench}.
We use GPT-5.4-mini for model-based judging.
The reported \emph{Acc} is the unweighted mean over these seven benchmarks.

\subsection{Main Results}
\label{sec:main-results}

\begin{table}[!t]
\centering
\caption{\textbf{Main results on seven benchmarks.}
Base, SFT, GRPO, answer-hint OPSD, and \method are evaluated across three Qwen3-VL and three Qwen3.5 scales.
Acc. is the unweighted mean of the seven benchmarks.
\method provides a strong reward-free result and improves the corresponding base model at every scale.
The highest score in each column within each model block is shown in bold.}
\label{tab:main}
\resizebox{\columnwidth}{!}{
\begin{tabular}{ll|ccccccc|c}
\toprule
\textbf{Model} & \textbf{Method} & \textbf{BLINK} & \textbf{MMStar} & \textbf{V$^*$} & \textbf{MathVista} & \textbf{HR4K} & \textbf{HR8K} & \textbf{HalluB} & \textbf{Acc.} \\
\midrule
\multirow{5}{*}{Qwen3-VL-2B}
& Base & 54.55 & 57.00 & 73.82 & 62.40 & 71.12 & 67.62 & 52.32 & 62.69 \\
& SFT & 50.92 & 58.40 & 70.68 & 61.30 & 71.62 & 68.50 & 50.63 & 61.72 \\
& GRPO & 55.29 & 62.13 & 74.87 & 64.90 & 76.62 & 71.50 & \textbf{53.84} & 65.59 \\
& OPSD & 53.87 & 60.67 & 73.30 & 63.20 & 72.88 & 71.00 & 50.53 & 63.63 \\
& \method{} (ours) & \textbf{57.92} & \textbf{63.67} & \textbf{78.53} & \textbf{68.50} & \textbf{77.62} & \textbf{72.75} & 53.41 & \textbf{67.49} \gain{4.80} \\
\midrule
\multirow{5}{*}{Qwen3-VL-4B}
& Base & \textbf{67.18} & 68.93 & 80.63 & 73.90 & 79.88 & 73.75 & 54.81 & 71.30 \\
& SFT & 66.02 & 68.60 & 82.72 & 75.10 & 78.62 & 74.12 & 52.26 & 71.06 \\
& GRPO & 65.70 & \textbf{69.80} & \textbf{83.77} & \textbf{75.60} & \textbf{81.12} & 71.37 & 60.14 & 72.50 \\
& OPSD & 64.97 & 68.73 & \textbf{83.77} & 74.10 & 78.12 & 75.63 & 54.50 & 71.40 \\
& \method{} (ours) & 66.86 & 68.73 & \textbf{83.77} & 74.90 & 80.50 & \textbf{77.00} & \textbf{60.37} & \textbf{73.16} \gain{1.86} \\
\midrule
\multirow{5}{*}{Qwen3-VL-8B}
& Base & 69.65 & 70.67 & 82.72 & 76.50 & 77.38 & 71.12 & 59.51 & 72.51 \\
& SFT & 66.33 & 72.00 & 86.39 & 75.80 & 79.50 & 75.62 & 60.27 & 73.70 \\
& GRPO & 68.65 & 73.40 & 86.39 & \textbf{79.30} & 78.75 & 74.87 & \textbf{60.83} & 74.60 \\
& OPSD & 67.81 & 70.20 & 85.34 & 77.30 & 81.12 & 74.62 & 59.64 & 73.72 \\
& \method{} (ours) & \textbf{70.38} & \textbf{74.07} & \textbf{87.43} & 77.90 & \textbf{84.12} & \textbf{79.25} & 60.69 & \textbf{76.26} \gain{3.75} \\
\midrule
\multirow{5}{*}{Qwen3.5-2B}
& Base & 59.02 & 67.67 & 83.25 & 74.10 & 73.75 & 72.25 & 50.22 & 68.61 \\
& SFT & 59.65 & 67.87 & 79.06 & 73.00 & 76.25 & 72.25 & 52.67 & 68.68 \\
& GRPO & 64.07 & 66.47 & \textbf{86.39} & 72.80 & 76.38 & 73.50 & 51.28 & 70.13 \\
& OPSD & 58.76 & 66.40 & 79.06 & 71.00 & 76.62 & 75.88 & 47.89 & 67.94 \\
& \method{} (ours) & \textbf{66.44} & \textbf{69.53} & 83.25 & \textbf{75.90} & \textbf{78.75} & \textbf{77.50} & \textbf{56.47} & \textbf{72.55} \gain{3.94} \\
\midrule
\multirow{5}{*}{Qwen3.5-4B}
& Base & 64.97 & 73.00 & 82.20 & 81.70 & 81.38 & 74.00 & 60.36 & 73.94 \\
& SFT & 65.39 & 73.00 & 86.39 & 81.30 & 83.50 & 77.50 & 61.66 & 75.54 \\
& GRPO & 65.75 & 73.60 & \textbf{86.91} & 79.10 & \textbf{86.88} & 73.87 & 58.67 & 74.97 \\
& OPSD & 64.44 & 71.40 & 84.29 & 77.60 & 82.75 & 78.75 & 58.22 & 73.92 \\
& \method{} (ours) & \textbf{66.70} & \textbf{75.00} & 84.82 & \textbf{82.20} & 85.38 & \textbf{79.38} & \textbf{63.95} & \textbf{76.77} \gain{2.83} \\
\midrule
\multirow{5}{*}{Qwen3.5-9B}
& Base & 66.70 & 74.60 & 83.25 & 81.60 & 82.38 & 77.38 & 58.89 & 74.97 \\
& SFT & 67.02 & 75.60 & 84.29 & 81.60 & 82.12 & 78.25 & 60.56 & 75.63 \\
& GRPO & 68.86 & 75.93 & 86.39 & 81.50 & 81.62 & 79.38 & 60.97 & 76.38 \\
& OPSD & 67.65 & 72.27 & \textbf{87.96} & 78.80 & 81.50 & 78.12 & 56.80 & 74.73 \\
& \method{} (ours) & \textbf{72.23} & \textbf{78.87} & 85.86 & \textbf{85.00} & \textbf{87.00} & \textbf{81.50} & \textbf{64.19} & \textbf{79.24} \gain{4.27} \\
\bottomrule
\end{tabular}
}

\vspace{-0.3cm}
\end{table}

Table~\ref{tab:main} consolidates the primary comparison across various Qwen model families and scales.

\textbf{A single reward-free objective delivers gains across all six scales.}
\method raises the corresponding base-model aggregate at every scale from Qwen3-VL-2B/4B/8B to Qwen3.5-2B/4B/9B, with gains between $+1.86\%$ and $+4.80\%$.
These results use neither cached target responses nor a verifiable reward, isolating input-conditioned self-distillation as the training signal.

\textbf{\method improves over answer-hint OPSD across visual capabilities.}
On Qwen3-VL-2B, \method improves every benchmark over OPSD.
This yields a $+3.86\%$ aggregate gain, with improvements spanning general perception, visual mathematics, fine-grained and high-resolution perception, and hallucination rather than being driven by a single benchmark.
Since OPSD supplies the teacher with the reference answer, this comparison supports visual conditioning as an alternative source of informative targets across these capabilities.

\textbf{Input-conditioned asymmetry scales across both model families.}
The gains over answer-hint OPSD persist at the largest tested scale in each family.
This extends the result from smaller models to stronger starting models, showing that the contrast remains useful as base capabilities increase within the tested families.
Matched visual predictions therefore provide effective self-distillation without answer or evidence hints.

\subsection{Generalization Across Training Datasets}
\label{sec:vision-opd-generalization}

The main results train \method on ViRL39K.
To test transfer across training corpora, we retrain \method on Vision-OPD6K using only its image--question pairs, without consuming bounding-box annotations.
Table~\ref{tab:vision-opd-comparison} compares Qwen3.5-4B and Qwen3.5-9B with Vision-OPD.

\begin{table*}[t]
\centering
\caption{\textbf{Generalization to Vision-OPD6K.} \method uses image--question pairs only; Vision-OPD rows use released checkpoints trained on the full corpus. Acc. averages the seven shared benchmarks, Avg.~(9) adds MME-RealWorld CN/EN, and bold marks the highest score per model block.}
\label{tab:vision-opd-comparison}
\resizebox{\textwidth}{!}{
\begin{tabular}{ll|ccccccc|c|cc|c}
\toprule
\textbf{Model} & \textbf{Method} & \textbf{BLINK} & \textbf{MMStar} & \textbf{V$^*$} & \textbf{MathVista} & \textbf{HR4K} & \textbf{HR8K} & \textbf{HalluB} & \textbf{Acc.} & \textbf{MME-RW CN} & \textbf{MME-RW EN} & \textbf{Avg.~(9)} \\
\midrule
\multirow{3}{*}{Qwen3.5-4B}
& Base & 64.97 & 73.00 & 82.20 & 81.70 & 81.38 & 74.00 & 60.36 & 73.94 & 64.20 & 62.90 & 71.63 \\
& Vision-OPD & 60.97 & 71.47 & \textbf{87.43} & 79.10 & 82.62 & 79.12 & 58.89 & 74.23 & \textbf{69.21} & \textbf{70.44} & 73.25 \\
& \method{} (ours) & \textbf{67.65} & \textbf{74.73} & 84.29 & \textbf{82.70} & \textbf{85.88} & \textbf{81.12} & \textbf{62.35} & \textbf{76.96} & 68.62 & 69.51 & \textbf{75.21} \\
\midrule
\multirow{3}{*}{Qwen3.5-9B}
& Base & 66.70 & 74.60 & 83.25 & 81.60 & 82.38 & 77.38 & 58.89 & 74.97 & 67.47 & 69.58 & 73.54 \\
& Vision-OPD & 66.23 & 75.33 & \textbf{92.15} & 82.60 & 85.50 & \textbf{85.38} & 59.44 & 78.09 & \textbf{71.17} & 72.20 & 76.67 \\
& \method{} (ours) & \textbf{69.81} & \textbf{77.47} & 90.58 & \textbf{83.50} & \textbf{88.25} & 83.62 & \textbf{61.57} & \textbf{79.26} & 68.35 & \textbf{72.54} & \textbf{77.30} \\
\bottomrule
\end{tabular}
}
\vspace{-0.3cm}
\end{table*}

\textbf{\method transfers beyond ViRL39K without supervision.}
In Table~\ref{tab:vision-opd-comparison}, \method reaches $76.96\%$ and $79.26\%$ aggregate accuracy on Qwen3.5-4B and Qwen3.5-9B, improving their corresponding base models by $3.02\%$ and $4.29\%$.
At both scales, \method also exceeds the released Vision-OPD on both Acc. and Avg.~(9), using only image--question pairs from the 6K subset, while Vision-OPD uses bounding-box supervision on the full corpus.
These results establish transfer across training corpora and higher aggregate performance without external information.

\subsection{Effect of Plausibility Restriction}
\label{sec:beta-support}

\begin{figure*}[t]
\centering
\includegraphics[width=\textwidth]{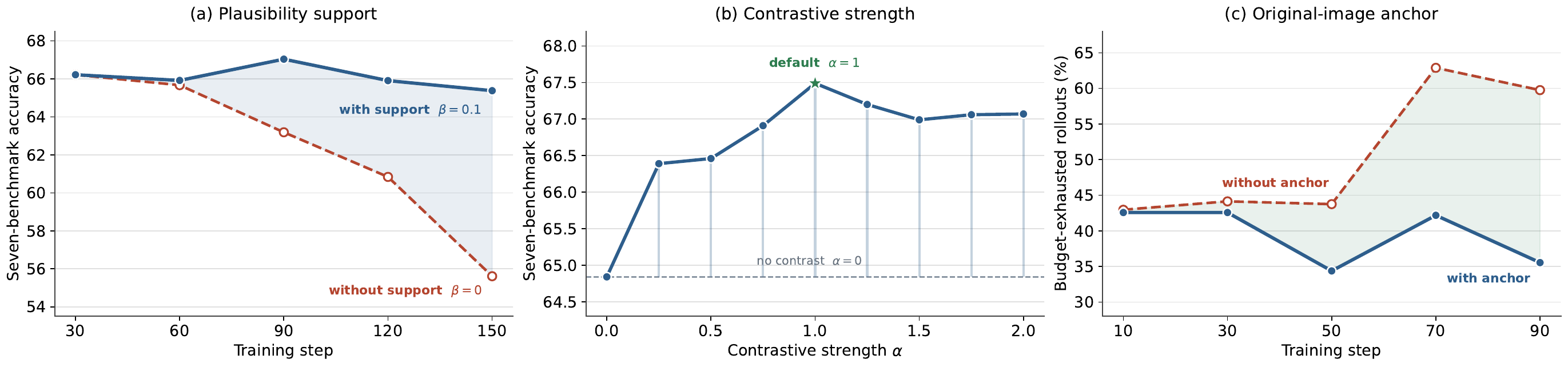}
\caption{\textbf{Target-shaping ablations on Qwen3-VL-2B.} \textbf{(a)} Seven-benchmark accuracy over extended training with and without plausibility support: the unrestricted target ($\beta=0$) degrades under continued self-distillation, while $\beta=0.1$ preserves accuracy. \textbf{(b)} Accuracy across contrast strengths: accuracy rises monotonically to its peak at $\alpha=1$ and stays within $0.6$ points of the peak for every strength in $[0.75, 2.0]$; $\alpha=1$ is the default. \textbf{(c)} Fraction of training rollouts that exhaust the generation budget without producing an answer, in a paired run differing only in the anchor coefficient: removing the anchor roughly doubles this fraction by the end of training.}
\label{fig:ablations}
\vspace{-0.3cm}
\end{figure*}

\textbf{Plausibility support protects accuracy during continued training.}
To examine its role, we compare training with plausibility support enabled ($\beta=0.1$) and removed ($\beta=0$). 
As shown in Figure~\ref{fig:ablations}(a), the two runs are indistinguishable through step $60$, after which the unsupported run degrades steadily, dropping to $55.6\%$ by step $150$, while the supported run stays near its starting accuracy ($65$--$67\%$). 
The resulting gap is $3.8$ points at step $90$ and $9.8$ points at step $150$.
Without this restriction, contrast can promote tokens that are unlikely under the original image simply because they are even less likely under the erased image. 
Plausibility support excludes these candidates from the training target, keeping repeated self-distillation focused on plausible predictions.

\subsection{Effect of Contrastive Strength}
\label{sec:alpha-ablation}

\textbf{\method is effective across a broad range of contrast strengths.}
To assess sensitivity to contrast strength, we vary $\alpha$ while keeping all other settings fixed.
The $\alpha=0$ control retains original-image self-distillation and plausibility support but removes contrastive reweighting.
As shown in Figure~\ref{fig:ablations}(b), accuracy increases from $64.84\%$ without contrast to $67.49\%$ at $\alpha=1$; even the smallest tested strength ($\alpha=0.25$) recovers $1.55$ points, and all tested strengths in $[0.75, 2.0]$ remain within $0.58\%$ of the peak.
This broad effective range indicates that the benefit comes from incorporating visual contrast rather than precisely tuning its strength.
We use $\alpha=1$ throughout the main experiments.

\paragraph{The original-image anchor improves rollout completion.}
We examine how the original-image anchor affects answer completion during training.
In paired runs, we remove the standalone original-image log-probability term while retaining visual contrast and plausibility support.
As shown in Figure~\ref{fig:ablations}(c), removing the anchor roughly doubles the fraction of rollouts that exhaust the generation budget without producing an answer by the end of training.
Table~\ref{tab:anchor-ablation} reports comparable seven-benchmark accuracy.
The anchor therefore complements plausibility restriction: the mask selects admissible tokens, while retaining original-image preferences helps rollouts complete their answers.

\begin{table}[t]
\centering
\caption{\textbf{Original-image anchor ablation on Qwen3-VL-2B.} No anchor removes the original-image log-probability term but retains contrast and plausibility support. Acc. averages seven benchmarks. Accuracy is comparable; Figure~\ref{fig:ablations}(c) shows the rollout-completion benefit.}
\label{tab:anchor-ablation}
\resizebox{0.95\columnwidth}{!}{%
\begin{tabular}{l|ccccccc|c}
\toprule
Config & BLINK & MMStar & V$^*$ & MathVista & HR4K & HR8K & HalluB & Acc. \\
\midrule
No anchor & 58.71 & 62.93 & 75.92 & 67.00 & 76.38 & 73.12 & 53.43 & 66.78 \\
With anchor (\method) & 57.92 & 63.67 & 78.53 & 68.50 & 77.62 & 72.75 & 53.41 & 67.49 \\
\bottomrule
\end{tabular}%
}
\vspace{-0.3cm}
\end{table}

\subsection{Effect of Distillation Divergence}
\label{sec:divergence-direction}

\textbf{Forward KL directly transfers the target's visual preferences.}
To isolate how the shaped target supervises the student, we keep target construction unchanged and vary the distillation divergence.
Table~\ref{tab:divergence-direction} shows comparable accuracy for forward KL and JSD.
Forward KL exceeds reverse KL by $3.06\%$ in aggregate accuracy, while differing from JSD by $0.17\%$.
Forward KL directly reinforces tokens promoted by the visual contrast, including those initially assigned low student probability.
Reverse KL weights discrepancies by the student's current probabilities, which can weaken this correction; this distinction motivates forward KL as our default.

\begin{table}[t]
\centering
\caption{\textbf{Distillation-divergence ablation on Qwen3-VL-2B.} Forward KL and JSD preserve the contrast-shaped target more effectively than reverse KL. }
\label{tab:divergence-direction}
\resizebox{0.9\columnwidth}{!}{%
\begin{tabular}{l|ccccccc|c}
\toprule
Divergence & BLINK & MMStar & V$^*$ & MathVista & HR4K & HR8K & HalluB & Acc. \\
\midrule
forward KL & \textbf{57.92} & \textbf{63.67} & \textbf{78.53} & 68.50 & \textbf{77.62} & 72.75 & 53.41 & \textbf{67.49} \\
JSD & 56.92 & 61.73 & \textbf{78.53} & \textbf{69.20} & 76.50 & \textbf{73.75} & \textbf{54.61} & 67.32 \\
reverse KL & 57.50 & 60.13 & 75.92 & 62.60 & 73.12 & 70.25 & 51.47 & 64.43 \\
\bottomrule
\end{tabular}
}

\vspace{-0.3cm}
\end{table}

\begin{table}[t]
\centering
\caption{\textbf{Control-image construction on Qwen3-VL-2B.} Full-removal controls behave similarly, while blur preserves coarse image content and trails. Natural references use other samples' images.
}
\label{tab:control-image-ablation}
\resizebox{0.9\columnwidth}{!}{%
\begin{tabular}{l|ccccccc|c}
\toprule
Config & BLINK & MMStar & V$^*$ & MathVista & HR4K & HR8K & HalluB & Acc. \\
\midrule
Gaussian noise & \textbf{58.50} & \textbf{63.80} & 76.44 & 64.50 & 76.63 & \textbf{73.12} & 53.04 & 66.58 \\
Black (ours) & 57.92 & 63.67 & \textbf{78.53} & \textbf{68.50} & \textbf{77.62} & 72.75 & 53.41 & \textbf{67.49} \\
Unrelated natural image & 56.86 & 63.60 & 75.92 & 66.20 & 77.50 & 72.62 & 53.92 & 66.66 \\
Gaussian blur & 55.18 & 60.87 & 75.92 & 61.80 & 74.87 & 69.38 & 49.98 & 64.00 \\
No image & 56.65 & 62.73 & 76.44 & 62.70 & 76.50 & 71.75 & \textbf{53.95} & 65.82 \\
\bottomrule
\end{tabular}%
}

\vspace{-0.3cm}
\end{table}

\paragraph{Content removal is the key property of the reference.}
To determine what makes a reference effective, we compare black, Gaussian-noise, blurred, and no-image controls on Qwen3-VL-2B.
These variants distinguish removing instance-specific content from merely degrading image appearance.
Table~\ref{tab:control-image-ablation} shows that black, noise, and no-image controls all outperform blur, with black improving aggregate accuracy by $3.49\%$ and MathVista by $6.70\%$ over blur.
We additionally replace black controls with shuffled images drawn from other training samples, keeping the training recipe unchanged.
This comparisons yield comparable aggregate accuracy with natural-image and black controls, suggesting that the benefit does not depend on a black reference specifically.
Together, these results support removing the current instance's evidence from the reference rather than retaining a coarse view of the same scene.
We therefore use black images as a simple deterministic reference throughout the main experiments.

\paragraph{\method maintains an accuracy advantage over OPSD throughout training.}
\label{sec:steps-curve}

\begin{figure}[!t]
  \centering
  \includegraphics[width=\textwidth]{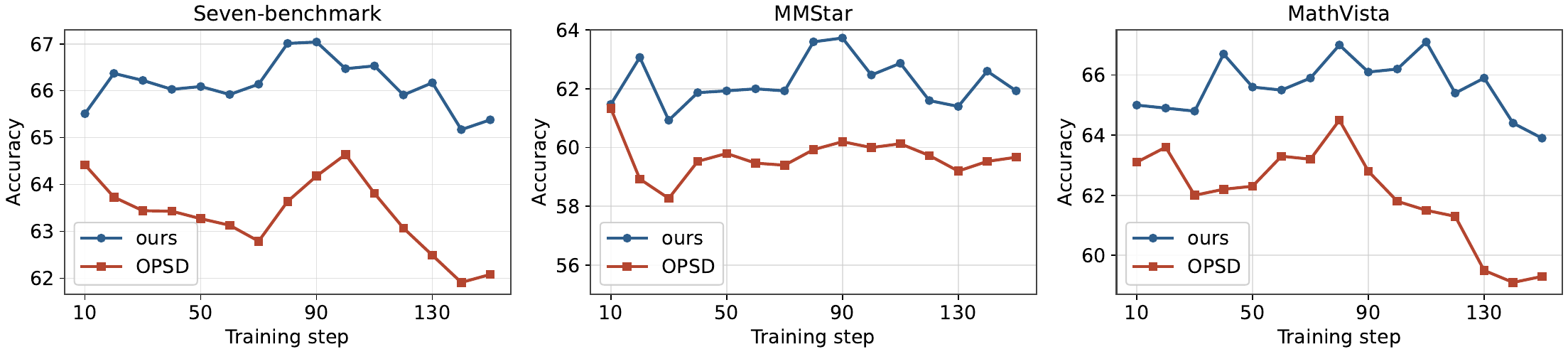}
  \caption{\textbf{Training dynamics of \method and OPSD on Qwen3-VL-2B.} Each panel tracks accuracy over training for the two runs: seven-benchmark aggregate, MMStar, and MathVista. \method leads on all three metrics throughout the evaluated training trajectory.}
  \label{fig:step-curve}
  
\vspace{-0.3cm}
\end{figure}

To examine whether the gains persist as training proceeds, we compare \method and answer-hint OPSD at matched training steps on Qwen3-VL-2B.
We track seven-benchmark mean accuracy, MMStar, and MathVista to examine aggregate performance, general perception, and visual mathematics.
As shown in Figure~\ref{fig:step-curve}, \method consistently leads on the aggregate, with an average absolute improvement of $2.80\%$ across the evaluated steps.
The advantage also holds on MMStar and MathVista throughout the measured trajectory, showing that visual contrast provides an accuracy advantage during both early learning and continued optimization.

\noindent\begin{minipage}{\linewidth}
\paragraph{\method benefits from visual-contrast training and teacher updates.}\mbox{}\par
\setlength{\intextsep}{0pt}
\setlength{\columnsep}{14pt}
\begin{wrapfigure}{r}{0.46\linewidth}
\centering
\includegraphics[width=\linewidth]{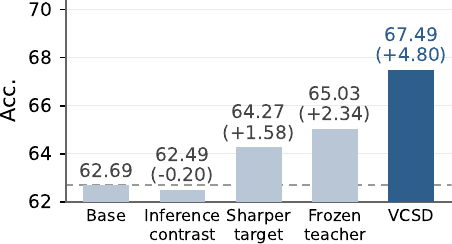}
\setlength{\abovecaptionskip}{4pt}
\caption{\textbf{Sources of \method's gains on Qwen3-VL-2B.} Bars report seven-benchmark mean accuracy. Parentheses show absolute changes from the base model, whose accuracy is marked by the dashed line.}
\label{fig:mechanism-ladder}
\end{wrapfigure}
\noindent
We investigate how visual-contrast targets and teacher updates contribute to \method's gains.
On Qwen3-VL-2B, we compare three controls: inference-only contrast applies the same contrast and plausibility mask at decoding without training; sharpening-only training doubles the original-image logits while removing the control branch; and the frozen-teacher variant fixes teacher parameters while retaining contrast and student training.
As shown in Figure~\ref{fig:mechanism-ladder}, \method improves accuracy by $4.80\%$ over the base, compared with $-0.20\%$ for inference-only contrast; both training controls also yield smaller gains.
These results support learning visually directed target changes beyond generic sharpening, with teacher updates contributing to the improvement.
\par
\end{minipage}\par


\section{Conclusion}
\label{sec:conclusion}
We introduced \method, which constructs OPSD target asymmetry from matched original and content-erased image conditions.
Their token-wise log-probability contrast sharpens the teacher's original-image distribution within plausible support, producing a full-distribution target without an external teacher, privileged answers, or external visual-evidence annotations.
Across six Qwen3-VL and Qwen3.5 scales, \method improves the corresponding base models and matched OPSD baselines on ViRL39K, with gains over the base models transferring to Vision-OPD6K.
Together, the consistent gains and target-design analyses establish conditional prediction changes as effective self-distillation supervision.
At deployment, resulting model retains backbone's standard single-pass inference path.


\clearpage

\bibliography{iclr2027/iclr2027_conference}

\newpage
\appendix
\section*{Appendix}
\section{Experimental Details}
\label{sec:appendix-repro}

\subsection{Baseline Implementations}
\textbf{GRPO.}
The reward is a rule-based verifiable check with no reward model and no KL term in the reward: the predicted answer is extracted from the response and matched against the reference.
Rollouts sample $n=8$ responses per prompt at temperature $1.0$ with a maximum response length of $1024$.
GRPO uses its own tuned hyperparameters at every scale (learning rate $1\times10^{-6}$, PPO mini-batch $8$, clip range $0.2$--$0.28$, $32$ prompts per step) with no per-scale hyperparameter search.

\textbf{SFT.}
The teacher cache is generated by the same backbone at each scale, sampling the fixed-seed $6{,}000$-prompt subset at temperature $1.0$ under the non-thinking chat template, whose generation prompt closes the reasoning block so that targets are direct answers.

\textbf{\method.}
Unless stated otherwise: $\alpha=1$, $\beta=0.1$, $T_{\mathrm{KD}}=2$, EMA rate $\rho=0.05$, uniform response-position weights, learning rate $2\times10^{-6}$, batch $32$ prompts with $n=8$ rollouts, $10$ warmup steps, and a fixed 90-step budget; the full-vocabulary forward-KL objective is optimized with AdamW.
Every training run in the paper uses a single 8-GPU node.
Measured wall clock for a 90-step \method run on Qwen3-VL-2B is about $2.0$ hours.

\subsection{Variance across Independent Training Seeds}
\label{sec:appendix-variance}
Table~\ref{tab:seed-variance} reports all settings trained with multiple independent seeds. Every checkpoint is evaluated under the same protocol, and \method{} outperforms GRPO, which in turn outperforms OPSD, for every seed.

\begin{table}[h]
\centering
\caption{\textbf{Variance across independent training seeds on Qwen3-VL-2B.} Each row reports the mean and standard deviation of the seven-benchmark accuracy over three independent training seeds.}
\label{tab:seed-variance}
\begin{tabular}{ccc}
\toprule
\textbf{\method{}} & \textbf{GRPO} & \textbf{OPSD} \\
\midrule
$67.49\pm0.51$ & $65.69\pm0.06$ & $63.63\pm0.77$ \\
\bottomrule
\end{tabular}
\end{table}



\section{Supplementary Experiments and Analyses}
\label{sec:appendix-analyses}

\subsection{Vision-OPD Native Evaluation}
\label{sec:appendix-vopd-native}

\begin{table*}[t]
\centering
\caption{\textbf{Vision-OPD native evaluation on Qwen3.5-4B.} All rows use non-thinking inference and the official \texttt{gpt-oss-120b} judge. \method and its bbox-crop variant train on Vision-OPD6K; the released Vision-OPD checkpoint uses the full corpus. Sup. denotes bounding-box supervision: the combined variant uses annotated crops for the teacher view together with visual contrast. Avg. is the unweighted mean of the six displayed benchmarks, and bold marks column maxima. The combined variant has the highest mean, supporting the complementarity of crop supervision and contrast.}
\label{tab:vopd6k-native}
\resizebox{\textwidth}{!}{
\begin{tabular}{llc|cccccc|c}
\toprule
\textbf{Model} & \textbf{Method} & \textbf{Sup.} & \textbf{V$^*$} & \textbf{ZoomBench} & \textbf{HR8K} & \textbf{HR4K} & \textbf{MME-RW CN} & \textbf{MME-RW EN} & \textbf{Avg.} \\
\midrule
\multirow{4}{*}{Qwen3.5-4B}
& Base & -- & 81.68 & 48.52 & 80.38 & 84.25 & 62.57 & 63.31 & 70.12 \\
& Vision-OPD (official) & bbox & \textbf{90.05} & 60.59 & 79.88 & 82.75 & \textbf{69.26} & 71.32 & 75.64 \\
& \method{} (ours) & none & 85.34 & 54.44 & 80.38 & \textbf{85.62} & 68.35 & \textbf{73.40} & 74.59 \\
\cmidrule(lr){2-10}
& \method{} + bbox-crop teacher & bbox & 89.53 & \textbf{62.25} & \textbf{82.12} & 84.50 & 68.36 & 72.02 & \textbf{76.46} \\
\bottomrule
\end{tabular}
}
\end{table*}

\textbf{Native evaluation separates localization and high-resolution strengths.}
On Qwen3.5-4B, \method without boxes provides the strongest HRBench4K and MME-RealWorld EN results, while the released Vision-OPD checkpoint leads on V$^*$ and MME-RealWorld CN.
Composing input contrast with the bbox-crop teacher reaches the strongest ZoomBench, HRBench8K, and native average, identifying the two supervision signals as complementary.
ZoomBench is measured under the separate native protocol of Table~\ref{tab:vopd6k-native} and is not included in Acc.
A matched no-contrast control (the Vision-OPD recipe retrained on the same data for the same 65 steps) scores $90.58$/$80.38$/$79.00$/$60.83$ on V$^*$/HRBench4K/HRBench8K/ZoomBench under the native protocol; adding the contrastive term contributes $+4.12$ on HRBench4K, $+3.12$ on HRBench8K, and $+1.42$ on ZoomBench, while retaining a V$^*$ score of $89.53$.

\subsection{Effect of Vocabulary Support}
\label{sec:support-ablation}

\begin{table}[t]
\centering
\caption{\textbf{Vocabulary support for the distillation loss.} We compare full-vocabulary distillation with student top-$K$ support and the union of the student's and target's top-$K$ candidates, renormalizing both distributions on the selected set. Columns correspond to Qwen3.5-4B trained on Vision-OPD6K and Qwen3-VL-4B trained on ViRL39K, each evaluated in one batch. Scores average the six main benchmarks excluding HRBench8K; this mean is specific to this table. Bold marks the best score per column. Union-support variants remain close to full-vocabulary distillation, with no shared best $K$ across the two settings.}
\label{tab:k-ablation}
\begin{tabular}{l|cc}
\toprule
\textbf{Support} & \textbf{Qwen3.5-4B} & \textbf{Qwen3-VL-4B} \\
\midrule
full vocabulary & 76.27 & 72.56 \\
student top-10 & 75.35 & 72.66 \\
student $\cup$ target top-10 & 76.24 & \textbf{73.23} \\
student $\cup$ target top-50 & \textbf{76.94} & 72.53 \\
student $\cup$ target top-100 & 75.89 & 72.95 \\
\bottomrule
\end{tabular}
\end{table}

\textbf{The objective is insensitive to the distillation support.}
We compare the full-vocabulary objective against variants that restrict the support at each position to the top-$K$ tokens of the student, or of the union of student and target, renormalizing both distributions on that set.
At $K{=}10$ the union variant matches full-vocabulary distillation across five backbones: $73.64$ versus $73.16$ on Qwen3-VL-4B and $76.91$ versus $76.77$ on Qwen3.5-4B (ViRL39K), $76.84$ versus $76.96$ on Qwen3.5-4B and $78.44$ versus $79.26$ on Qwen3.5-9B (Vision-OPD6K), $74.80$ versus $74.27$ on Qwen3-VL-8B (Vision-OPD6K), and $65.98$ versus $67.49$ on Qwen3-VL-2B (ViRL39K).
Table~\ref{tab:k-ablation} sweeps $K$: including the target's candidates adds $+0.89$ over student-only top-10 at the same $K$ on Qwen3.5-4B, and the Qwen3.5-4B ladder peaks at $K{=}50$.
The preferred $K$ shifts across datasets while the tested union-support settings remain close to full-vocabulary distillation; we use the full-vocabulary objective as the default and retain top-$K$ union support as the memory-efficient option, since it avoids materializing full-vocabulary teacher distributions.

\subsection{Effect of Original-Image Anchor}
\label{sec:anchor-drift}

To isolate the role of the original-image anchor, we introduce an
anchor coefficient $\lambda$ into the target score:
\begin{equation}
a_t^{(\lambda)}(v)
=
\lambda\log p_{\phi,t}^{I}(v)
+\alpha\widetilde{\Delta}_t(v)
=
(\lambda+\alpha)\log p_{\phi,t}^{I}(v)
-\alpha\log p_{\phi,t}^{0}(v),
\quad v\in\mathcal S_t(\beta).
\end{equation}
The full method uses $\lambda=1$, recovering
Eq.~\ref{eq:equivalent-target-score}, whereas the no-anchor variant sets
$\lambda=0$ while retaining the same plausibility support. With
$\alpha=1$, its target score becomes
$\log p_{\phi,t}^{I}(v)-\log p_{\phi,t}^{0}(v)$.

\textbf{The anchor stabilizes rollout generation while leaving aggregate accuracy unchanged.}
Table~\ref{tab:anchor-ablation} isolates the anchor from contrastive shaping and shows that the aggregate accuracy is comparable across the two configurations.
In paired runs differing only in the anchor coefficient, removing the anchor raises the fraction of rollouts that exhaust the generation budget without producing an answer from about one third to about three fifths by the end of training, on both seeds, while validation accuracy and format compliance are unchanged.
By keeping the shaped target close to the original prediction, the anchor keeps rollouts terminating and thereby serves as a training-time stability component.

\subsection{Qualitative Analysis}
\label{sec:qualitative-analysis}

MathVista provides a concrete setting for examining whether the aggregate improvement reflects more accurate use of image evidence.
\textbf{The case study localizes the decisive correction to visual counting.}
Figure~\ref{fig:casestudy} presents a base-ten block question whose answer depends on the number of thousand-cubes.
The base model misinterprets the place-value blocks and predicts $6169$, whereas OPSD overestimates the number of thousand-cubes and predicts $8519$.
\method correctly identifies seven thousand-cubes (six in the top row and one below), producing the correct answer of $7519$.
Once this image-grounded quantity is recovered, the remaining place-value calculation is direct.
This example is consistent with the conditioning contrast shifting the self-distillation target toward candidates supported by instance-specific visual content.

\begin{figure}[t]
\centering
\begin{minipage}[t]{0.34\linewidth}
\vspace{0pt}\centering
\includegraphics[width=\linewidth]{figures/casestudy_mathvista_163.png}\\[5pt]
{\scriptsize\textit{Q: What number is shown?}\\[3pt]\textit{Ground truth: $7519$}}
\end{minipage}
\hfill
{\scriptsize
\begin{minipage}[t]{0.62\linewidth}
\vspace{0pt}
\textbf{Qwen3.5-2B Base}\hfill predicts \textbf{6169}~\textcolor{red}{(wrong)}\\
Top row: $6$ large cubes $\to 6{,}000$. \textcolor{red}{Middle: one medium cube, representing $100$} $\to +100$. Bottom: $5$ flats, $1$ rod, $9$ units $\to 69$. Total $=6{,}169$.
\\[4pt]
\textbf{OPSD}\hfill predicts \textbf{8519}~\textcolor{red}{(wrong)}\\
Large cubes (thousands): \textcolor{red}{$7$ at the top and $1$ below, so $7+1=8$ thousands} $\to 8{,}000$. Then $5$ flats ($500$), $1$ rod ($10$), $9$ units ($9$). Total $=8{,}519$.
\\[4pt]
\textbf{\method{} (ours)}\hfill predicts \textbf{7519}~\textcolor{green!55!black}{(correct)}\\
The top row has \textcolor{green!55!black}{$6$ large cubes and $1$ more below $\to 7$ thousand-cubes ($7{,}000$)}. Plus \textcolor{green!55!black}{$5$ flats ($500$), $1$ rod ($10$), and $9$ units ($9$)}. Total $=7{,}519$.
\end{minipage}
}
\caption{\textbf{Visual counting in a MathVista example with Qwen3.5-2B.} The image and question appear on the left; response excerpts from the base model, OPSD, and \method appear on the right. Red highlights counting or place-value errors, and green highlights image-consistent quantities. \method correctly identifies the thousand-cubes and combines them with the remaining place-value blocks to obtain the ground-truth answer.}
\label{fig:casestudy}
\end{figure}

\begin{figure*}[t]
\centering
\includegraphics[width=\textwidth]{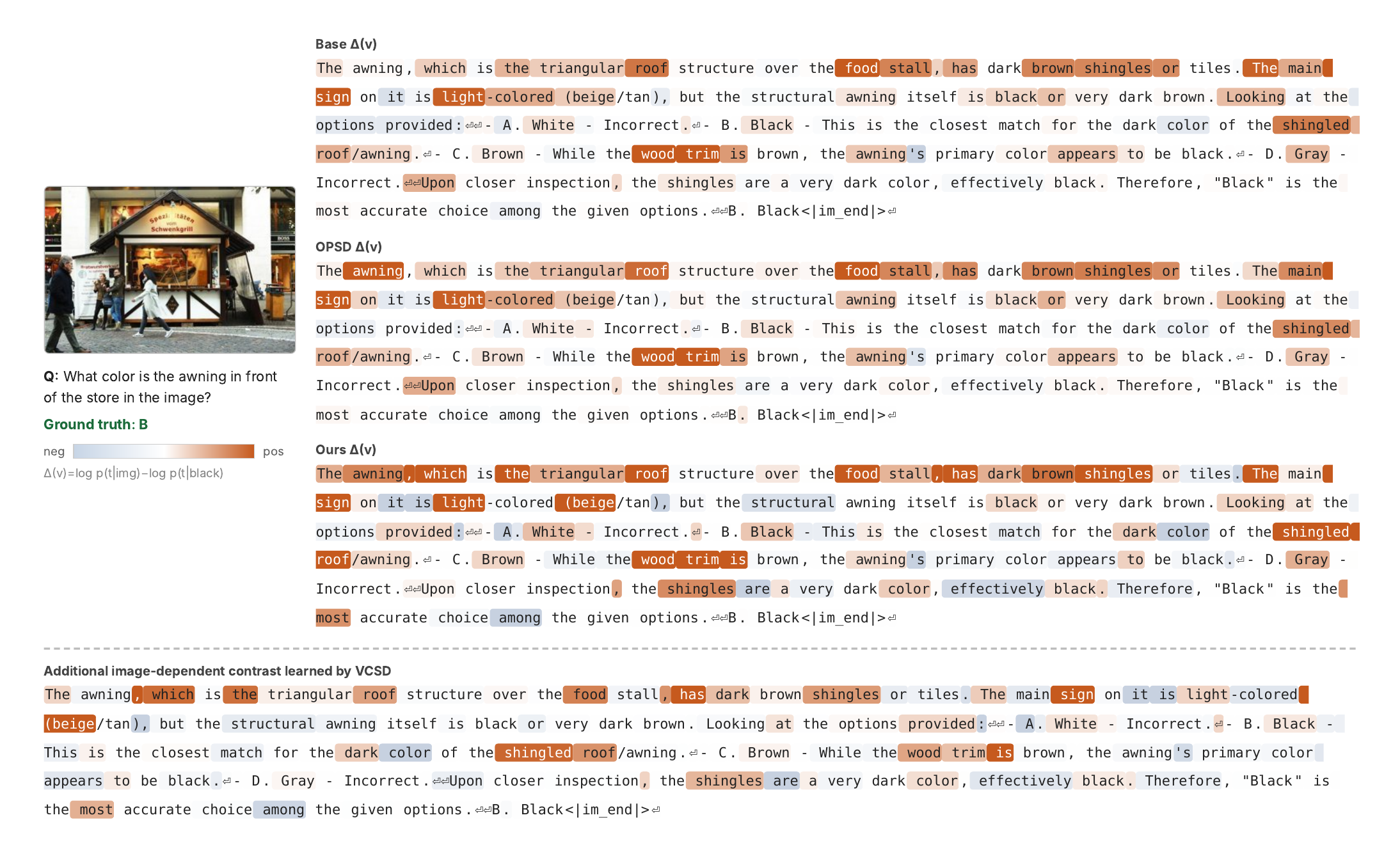}
\caption{\textbf{Token-level image-dependent contrast.} The base model, OPSD, and \method score the same response at matching prefixes under the displayed image and a black-image control. Token colors encode the log-probability difference $\Delta(v)=\log p(v\mid I)-\log p(v\mid I_{\mathrm{ctrl}})$: warmer colors indicate greater preference under the original image. The bottom row shows $\Delta_{\text{\method}}-\Delta_{\mathrm{OPSD}}$, isolating the change relative to OPSD. Stronger positive changes on visual terms such as \emph{roof}, \emph{shingles}, and \emph{beige} illustrate how the learned contrast varies across tokens.}
\label{fig:contrast-vis}
\end{figure*}

Figure~\ref{fig:contrast-vis} visualizes the learned image-dependent contrast score
$\Delta(v)=\log p(v\mid I)-\log p(v\mid I_{\mathrm{ctrl}})$
for the same generated response.
\textbf{\method concentrates additional contrast on image-grounded evidence.}
The base model and OPSD exhibit highly similar contrast patterns, whereas \method assigns stronger positive contrast to visual concepts such as \emph{roof}, \emph{shingles}, \emph{food}, \emph{beige}, and \emph{sign}, while suppressing less informative tokens.
The bottom row isolates the additional contrast learned over OPSD and shows that this change is localized to semantically meaningful visual concepts rather than distributed uniformly across the sequence.
This pattern supports the hypothesis that contrastive target shaping reallocates probability mass toward image-dependent tokens instead of merely amplifying existing language preferences.

\section{Theoretical Perspective}
\label{sec:theoretical-perspective}

The conditional log-ratio in Equation~\ref{eq:conditioning-contrast} admits two complementary interpretations. First, it can be viewed as a dense token-level reward measuring the predictive support attributable to instance-specific visual content. Second, it provides a controlled approximation to the conditional pointwise mutual information between a candidate token and the observed image, with the content-erased prediction serving as a surrogate for the prediction without instance-specific visual information.

\textbf{One-Step KL-Regularized Policy Update.}
Let the original-image teacher distribution restricted and renormalized over the plausibility support be

\begin{equation}
\bar p_{\phi,t}^{I}(v)
=
\frac{
\mathbf{1}\!\left[v\in\mathcal{S}_t(\beta)\right]
p_{\phi,t}^{I}(v)
}{
\sum_{u\in\mathcal{S}_t(\beta)}
p_{\phi,t}^{I}(u)
}.
\label{eq:support-normalized-original}
\end{equation}

We regard the conditioning contrast

\begin{equation}
r_t^{\mathrm{vis}}(v)
:=
\widetilde{\Delta}_t(v)
\label{eq:implicit-visual-reward}
\end{equation}

as an implicit visual-evidence reward.

\textbf{Remark 1.}
For each student-generated prefix, the contrast-shaped target in Equation~\ref{eq:contrast-shaped-target} is the unique solution to the following optimization problem:

\begin{equation}
q_t^\star
=
\arg\max_{q\in\Pi(\mathcal{S}_t(\beta))}
\left\{
\alpha
\mathbb{E}_{v\sim q}
\left[
r_t^{\mathrm{vis}}(v)
\right]
-
D_{\mathrm{KL}}
\left(
q
\,\Vert\,
\bar p_{\phi,t}^{I}
\right)
\right\}.
\label{eq:kl-regularized-policy-improvement}
\end{equation}

Here, $\Pi(\mathcal{S}_t(\beta))$ denotes the probability simplex over $\mathcal{S}_t(\beta)$. The closed-form solution is

\begin{equation}
q_t^\star(v)
=
\frac{
\bar p_{\phi,t}^{I}(v)
\exp\left(
\alpha r_t^{\mathrm{vis}}(v)
\right)
}{
\sum_{u\in\mathcal{S}_t(\beta)}
\bar p_{\phi,t}^{I}(u)
\exp\left(
\alpha r_t^{\mathrm{vis}}(u)
\right)
}.
\label{eq:closed-form-policy-improvement}
\end{equation}

Because the support-normalization constant in $\bar p_{\phi,t}^{I}$ cancels between the numerator and denominator, Equation~\ref{eq:closed-form-policy-improvement} is exactly equivalent to the target defined in Equation~\ref{eq:contrast-shaped-target}. The original-image prediction therefore acts as the reference policy, while the conditioning log-ratio provides the reward used to improve it. The plausibility restriction further imposes a hard support constraint, preventing tokens with large likelihood ratios but negligible original-image probability from dominating the improved target. A complete derivation is provided in Appendix~\ref{sec:kl-policy-proof}.

Here, ``one-step'' refers to a single closed-form update from the support-normalized original-image teacher distribution $\bar p_{\phi,t}^{I}$ to the locally improved target $q_t^\star$ at a fixed response prefix. It does not refer to a single gradient step of student optimization.

Importantly, this interpretation does not assume that the original-image policy was historically obtained by applying reinforcement learning to the control-image policy. Rather, the conditional log-ratio defines an implicit reward whose KL-regularized optimum coincides with our contrast-shaped target.

\textbf{Approximation to Conditional Pointwise Mutual Information.}
Let $H_t=(P,y_{<t})$ denote the textual context at step $t$. The conditional pointwise mutual information between token $v$ and image $I$, conditioned on $H_t$, takes the form

\begin{equation}
\operatorname{PMI}(v;I\mid H_t)
=
\log
\frac{
p(v\mid I,H_t)
}{
p(v\mid H_t)
}.
\label{eq:conditional-pmi}
\end{equation}

Our conditioning contrast instead takes the form

\begin{equation}
\Delta_t(v)
=
\log
\frac{
p_{\phi}(v\mid I,H_t)
}{
p_{\phi}(v\mid I_{\mathrm{ctrl}},H_t)
}.
\label{eq:controlled-conditional-ratio}
\end{equation}

When the content-erased prediction $p_{\phi}(v\mid I_{\mathrm{ctrl}},H_t)$ approximates the model's prediction without instance-specific visual information, it serves as a controlled surrogate for $p_{\phi}(v\mid H_t)$. Under this interpretation, $\Delta_t(v)$ is a contrastive approximation to conditional pointwise mutual information: it measures how much the observed image changes the token's log-likelihood beyond the content-independent prediction.
Positive values identify tokens whose likelihood increases in the presence of the observed visual content, whereas values near zero identify tokens whose likelihood is largely preserved after that content is removed. The guarded reward $r_t^{\mathrm{vis}}(v)=\widetilde{\Delta}_t(v)$ follows this interpretation for ordinary tokens, while designated termination tokens are excluded from contrastive adjustment for training stability.

This shows why the proposed objective is useful. Maximizing conditional mutual information favors predictions that remain informative about the observed image after accounting for the textual context, rather than predictions that can be explained primarily by language priors. Accordingly, upweighting tokens with large positive contrast encourages the student to preserve image-dependent evidence in its next-token distribution, while leaving visually insensitive tokens largely unchanged. Combined with the original-image teacher distribution as the reference policy, this yields a conservative form of visual grounding: the method amplifies visually supported distinctions only among tokens that the teacher already considers plausible, instead of rewarding arbitrary image-sensitive outputs. The resulting target therefore promotes stronger dependence on instance-specific visual evidence while retaining the fluency and semantic knowledge encoded by the teacher.

\section{Proof of Remark 1}
\label{sec:kl-policy-proof}

We prove Remark~1 from Section~\ref{sec:theoretical-perspective}. Fix a student-generated prefix $y_{<t}$, and abbreviate the plausibility support as $\mathcal{S}_t=\mathcal{S}_t(\beta)$. As in Equation~\ref{eq:support-normalized-original}, let

\begin{equation}
\bar p_{\phi,t}^{I}(v)
=
\frac{
\mathbf{1}\!\left[v\in\mathcal{S}_t\right]
p_{\phi,t}^{I}(v)
}{
\sum_{u\in\mathcal{S}_t}
p_{\phi,t}^{I}(u)
}
\label{eq:appendix-support-normalized-original}
\end{equation}

denote the original-image teacher distribution renormalized over $\mathcal{S}_t$, and, as in Equation~\ref{eq:implicit-visual-reward}, let

\begin{equation}
r_t^{\mathrm{vis}}(v)
=
\widetilde{\Delta}_t(v)
\label{eq:appendix-implicit-visual-reward}
\end{equation}

denote the implicit visual-evidence reward.

Consider the following optimization over $\Pi(\mathcal{S}_t)$, the probability simplex over $\mathcal{S}_t$:

\begin{equation}
\begin{aligned}
\max_{q\in\Pi(\mathcal{S}_t)}
\mathcal{J}_t(q)
:=
\alpha
\sum_{v\in\mathcal{S}_t}
q(v)r_t^{\mathrm{vis}}(v)
-
\sum_{v\in\mathcal{S}_t}
q(v)
\log
\frac{
q(v)
}{
\bar p_{\phi,t}^{I}(v)
}.
\end{aligned}
\label{eq:appendix-policy-objective}
\end{equation}

\textbf{Existence and uniqueness of the maximizer.}
Since the teacher distribution is a (temperature-scaled) softmax output, $p_{\phi,t}^{I}(v)>0$ for every $v\in\mathcal{V}$; hence, for any $\beta\in[0,1]$, $\bar p_{\phi,t}^{I}(v)>0$ for every $v\in\mathcal{S}_t$.
Under the convention $0\log 0=0$, the objective $\mathcal{J}_t$ is therefore finite and continuous on $\Pi(\mathcal{S}_t)$; since $\Pi(\mathcal{S}_t)$ is compact, a maximizer exists.
For uniqueness, decompose the objective as

\begin{equation}
\mathcal{J}_t(q)
=
\sum_{v\in\mathcal{S}_t}
q(v)
\left[
\alpha r_t^{\mathrm{vis}}(v)
+
\log \bar p_{\phi,t}^{I}(v)
\right]
-
\sum_{v\in\mathcal{S}_t}
q(v)\log q(v).
\label{eq:appendix-objective-decomposition}
\end{equation}

The first sum is linear in $q$, with finite coefficients because $\bar p_{\phi,t}^{I}$ is strictly positive on $\mathcal{S}_t$, while the remaining term $-\sum_{v\in\mathcal{S}_t}q(v)\log q(v)$ is the Shannon entropy of $q$, which is strictly concave on $\Pi(\mathcal{S}_t)$.
Hence $\mathcal{J}_t$ is strictly concave, and its maximizer is unique.

\textbf{Deriving the candidate solution.}
We first solve the problem subject only to the normalization constraint $\sum_{v\in\mathcal{S}_t}q(v)=1$, and verify afterwards that the omitted nonnegativity constraints $q(v)\geq 0$ are inactive at the resulting solution.
Introduce a Lagrange multiplier $\lambda$ for the normalization constraint:

\begin{equation}
\begin{aligned}
\mathcal{F}(q,\lambda)
=
\alpha
\sum_{v\in\mathcal{S}_t}
q(v)r_t^{\mathrm{vis}}(v)
-
\sum_{v\in\mathcal{S}_t}
q(v)
\log
\frac{
q(v)
}{
\bar p_{\phi,t}^{I}(v)
}
+
\lambda
\left(
\sum_{v\in\mathcal{S}_t}
q(v)-1
\right).
\end{aligned}
\label{eq:appendix-lagrangian}
\end{equation}

For every $v\in\mathcal{S}_t$, the stationarity condition is

\begin{equation}
\frac{\partial\mathcal{F}}{\partial q(v)}
=
\alpha r_t^{\mathrm{vis}}(v)
-
\log
\frac{
q(v)
}{
\bar p_{\phi,t}^{I}(v)
}
-
1
+
\lambda
=
0.
\label{eq:appendix-stationarity}
\end{equation}

Rearranging gives

\begin{equation}
\log
\frac{
q(v)
}{
\bar p_{\phi,t}^{I}(v)
}
=
\alpha r_t^{\mathrm{vis}}(v)
+
\lambda
-
1,
\label{eq:appendix-log-ratio-stationarity}
\end{equation}

and hence

\begin{equation}
q(v)
=
\bar p_{\phi,t}^{I}(v)
\exp\left(
\alpha r_t^{\mathrm{vis}}(v)
\right)
\exp(\lambda-1).
\label{eq:appendix-unnormalized-solution}
\end{equation}

The factor $\exp(\lambda-1)$ is shared across all candidates and is determined by normalization. Therefore,

\begin{equation}
q_t^\star(v)
=
\frac{
\bar p_{\phi,t}^{I}(v)
\exp\left(
\alpha r_t^{\mathrm{vis}}(v)
\right)
}{
\sum_{u\in\mathcal{S}_t}
\bar p_{\phi,t}^{I}(u)
\exp\left(
\alpha r_t^{\mathrm{vis}}(u)
\right)
}.
\label{eq:appendix-closed-form-policy}
\end{equation}

\textbf{Verifying global optimality.}
Every factor on the right-hand side of Equation~\ref{eq:appendix-closed-form-policy} is strictly positive, so $q_t^\star(v)>0$ for all $v\in\mathcal{S}_t$, and the nonnegativity constraints omitted above are strictly satisfied and hence inactive at $q_t^\star$.
Together with the stationarity condition in Equation~\ref{eq:appendix-stationarity} and primal feasibility, $q_t^\star$ therefore satisfies the Karush--Kuhn--Tucker (KKT) conditions of the problem in Equation~\ref{eq:appendix-policy-objective}, with zero multipliers on the inactive nonnegativity constraints.
Since the objective is concave and all constraints are affine, the KKT conditions are sufficient for global optimality, so $q_t^\star$ is the maximizer of Equation~\ref{eq:appendix-policy-objective}; by the strict concavity established above, it is the unique one.
Intuitively, no maximizer can place zero mass on any candidate in $\mathcal{S}_t$, since $\partial\mathcal{J}_t/\partial q(v)\to+\infty$ as $q(v)\to0^{+}$.

\textbf{Equivalence with the contrast-shaped target.}
For $v\in\mathcal{S}_t$, substituting

\begin{equation}
\bar p_{\phi,t}^{I}(v)
=
\frac{
p_{\phi,t}^{I}(v)
}{
\sum_{u\in\mathcal{S}_t}
p_{\phi,t}^{I}(u)
}
\label{eq:appendix-support-normalization}
\end{equation}

into Equation~\ref{eq:appendix-closed-form-policy}, the support-normalization constant appears in both the numerator and denominator and therefore cancels. We obtain

\begin{equation}
q_t^\star(v)
=
\frac{
\mathbf{1}\!\left[v\in\mathcal{S}_t\right]
p_{\phi,t}^{I}(v)
\exp\left(
\alpha\widetilde{\Delta}_t(v)
\right)
}{
\sum_{u\in\mathcal{S}_t}
p_{\phi,t}^{I}(u)
\exp\left(
\alpha\widetilde{\Delta}_t(u)
\right)
}.
\label{eq:appendix-recovered-target}
\end{equation}

This is exactly the contrast-shaped target in Equation~\ref{eq:contrast-shaped-target}, which proves the result.

For $\alpha>0$, an equivalent form of the objective is

\begin{equation}
q_t^\star
=
\arg\max_{q\in\Pi(\mathcal{S}_t)}
\left\{
\mathbb{E}_{v\sim q}
\left[
r_t^{\mathrm{vis}}(v)
\right]
-
\frac{1}{\alpha}
D_{\mathrm{KL}}
\left(
q
\,\Vert\,
\bar p_{\phi,t}^{I}
\right)
\right\}.
\label{eq:appendix-equivalent-policy-objective}
\end{equation}

Thus, increasing $\alpha$ assigns greater weight to the visual-evidence reward relative to the KL regularizer, while the limit $\alpha\to0$ recovers the support-normalized original-image teacher distribution.

\section{AI Use Statement}
\label{app:ai_use_statement}

We used generative AI tools to improve the manuscript's readability and linguistic clarity, assist with drafting and revising the discussion of experimental results, review the implementation of our method, and revise figure-generation code.

The authors reviewed the AI-assisted text and code-review suggestions.
We checked that manuscript revisions were consistent with our intended meaning and reported results, and assessed code suggestions against the corresponding implementation.
We take full responsibility for all content of this work, including its claims, theoretical arguments, experimental analysis, code, and figures.

\end{document}